\documentclass{ws-ijns}
\usepackage{url}
\usepackage{graphicx}
\usepackage[english]{babel}
\usepackage[super,sort,compress]{cite}
\usepackage{amsmath}
\usepackage{multirow}
\begin{document}

\catchline{0}{0}{2021}{}{}

\markboth{Binyi Wu}{Convolutional Neural Networks Quantization with Attention}

\title{Convolutional Neural Networks Quantization with Attention}

\author{Binyi Wu}
\address{Faculty of Electrical and Computer Engineering, Technische Universität Dresden, Helmholtzstr. 18,\\
Dresden, 01069, Germany\\
binyi.wu@tu-dresden.de, binyi.wu@infineon.com\\
www.tu-dresden.de}

\author{Bernd Waschneck}
\address{Division of Power and Sensor Systems, Infineon Technologies AG, Königsbrücker Str. 180,\\
Dresden, 01099, Germany\\
bernd.waschneck@infineon.com\\
www.infineon.com}

\author{Christian Georg Mayr}
\address{Centre for Tactile Internet with Human-in-the-loop (CeTI), Technische Universität Dresden, Helmholtzstr. 18,\\
Dresden, 01069, Germany\\
christian.mayr@tu-dresden.de\\
www.tu-dresden.de}

\maketitle

\begin{abstract}
It has been proven that, compared to using 32-bit floating-point numbers in the training phase, Deep Convolutional Neural Networks (DCNNs) can operate with low precision during inference, thereby saving memory space and power consumption. However, quantizing networks is always accompanied by an accuracy decrease. Here, we propose a method, double-stage Squeeze-and-Threshold (double-stage ST). It uses the attention mechanism to quantize networks and achieve state-of-art results. Using our method, the 3-bit model can achieve accuracy that exceeds the accuracy of the full-precision baseline model. The proposed double-stage ST activation quantization is easy to apply: inserting it before the convolution.
\end{abstract}


\begin{multicols}{2}

\section{Introduction}
To enable neural networks to function on edge devices, researchers have developed various optimization methods in the past few years, such as {quantization\cite{quantization}}, {pruning\cite{pruning1, pruning2}}, and {neural architecture search\cite{mobilenetv3}}. These methods optimize the network from different perspectives and are not mutually exclusive. Our work mainly studies network quantization.

Quantization not only saves storage costs by several or even tens of times, but it also reduces computing complexity by switching from floating-point to integer operations. Both optimizations brought by quantization are especially beneficial for embedded devices that run neural networks with limited memory and computing resources. Most of the current deep learning libraries, such as TensorFlow\cite{tensorflow} and {PyTorch\cite{pytorch}}, only support mature 16/8-bit quantization without loss of accuracy. However, for applications targetting extremely low costs, 16/8-bit quantization is still insufficient.

Previous work\cite{pact,lsq,reactnet} has shown that networks inference with low bit-width is feasible, and milestone progress has been made. On the other side, given the great success of the attention mechanism on CNNs, we found that it also helps to quantize the networks. Based on the previous work, {ST quantization\cite{stquantization}}, we further study the application of the attention mechanism in network quantization. We upgrade ST to double-stage ST, which is illustrated in Section 3. Section 4 demonstrates the experimental results. The conclusion part completes the paper. The proposed double-stage ST quantization overcomes the hardware-unfriendly-input-channel-wise quantization in ST and further improves the accuracy of low-precision networks. In particular, the accuracy of the 3-bit model exceeds the accuracy of the baseline model.

\section{Related Work}
For network optimization, there are quantization, pruning, and neural architecture search. Quantization optimizes the network at the lowest level, that is, the basic operation unit (the multiplication and accumulation operation). Pruning improves efficiency by cutting very weak or useless filters at the kernel level. NAS searches for the best network at the architectural level and combines different blocks to form an efficient network.

The quantization of the network starts with the quantization of weights. Ref.~\refcite{weightquantization} uses three values of weights +1, 0, and -1 to infer the network. It proves the feasibility of weight quantization. However, the performance benefit is restricted when only the weights are quantized.  Because the operation is still floating-point-based and the runtime memory used to store activations remains unchanged.  To further improve inference performance, activation also needs to be quantized. However, activation quantization is more problematic due to the usage of non-differentiable operators, which causes the gradient backpropagation to be erroneous.\cite{bnn} The commonly used solution is to approximate the quantization operation to {a straight-through estimator (STE)~\cite{ste}}.
In binarization, Ref.~\refcite{bnn} is the first to propose method of training Binary Neural Networks (BNNs). In the forward path, the $\mathrm{sign}$ function is employed for binarization, while in the backward path, STE is used. Binarization replaces arithmetic operations with bitwise operations, $\mathrm{XOR}$, which boosts efficiency significantly. However, it comes at the high expense of accuracy. Ref.~\refcite{bnn+} proposes a new binarization function, $\mathrm{SignSwish}$, as well as a regularization term to improve the convergence and generalization accuracy of binary networks. Researchers discovered that the limited representational capability of binary causes the decay. Therefore, subsequent research focuses on enhancing the binary network representation. Firstly, Ref.~\refcite{xnornet} created the scaling factors to scale the activation, which improves the top-1 accuracy by more than $16\%$ when compared to the previous state-of-the-art works. However, the introduced scaling factors have the same size as the activation and are of full precision. Therefore, in the later work Ref.~\refcite{xnornet++}, the scaling factors were reduced to be spatially and channel-wise. Unlike previous works of {Refs.~\refcite{xnornet,xnornet++}}, Ref.~\refcite{birealnet} chooses to strengthen the feature expression by connecting real activations to binary activations via a skip connection. In the end, it significantly strengthens the network expression ability, and the top-1 accuracy of the binary network was again greatly improved by $5\%$.
The previous work is mainly based on the existing network architecture, but they are not necessarily suitable for BNN. So afterward, researchers have successively studied the network architecture specific to BNN. With a dedicated BNN architecture, BinaryDenseNet~\cite{binarydenseNet} achieved another milestone and significantly exceeded all other 1-bit CNNs by more than $6\%$. In addition, Ref.~\refcite{meliusnet} demonstrated an architecture composed of $\mathrm{DenseBlock}$ and $\mathrm{ImprovementBlock}$ and proved such a network can improve feature capacity and feature quality. Whereas Ref.~\refcite{reactnet} observes that the performance of the binary network is sensitive to changes in the activation distribution. Therefore, they developed RSign and RPReLU from Sign and ReLU to explicitly learn distribution variations with almost no added cost. Because the binarization operation will cause the information loss in the activation forward and gradient backward, the performance gap between the full-precision model and the binary model is established. To overcome this problem, Ref.~\refcite{irnet} developed an information retention network, IR-Net, to maintain information in the forward activation and backward gradient.

For multi-bit quantization, Ref.~\refcite{dorefa} introduced DoReFa-Net to train multi-bit neural networks using gradients in non-full-precision. 
However, the ReLU non-linearity approximation in quantization causes the gradient mismatch. To address this, Ref.~\refcite{hwgq} offered a half-wave Gaussian quantizer (HWGQ) that approximates the quantizer using activation statistics. 
However, the multi-bit networks still suffer from accuracy degradation. To bridge the gap, Ref.~\refcite{pact} developed parameterized clipping activation (PACT), which has a parameter for learning the activation clipping value and is optimized by back-propagation. 
It is the first time that a 4-bit network can achieve accuracy comparable to full precision networks across a variety of popular models and datasets because to the power of gradient.
Whereas Ref.~\refcite{auxiliary} recommends using a full-precision auxiliary module to train the low-precision network, which updates the parameters with additional full-precision routes.
To address the non-differentiable problem of quantization, Ref.~\refcite{rq} proposed a differentiable quantization approach, which is done by turning continuous weights and activation distributions into categorical distributions using the quantization grid.
Afterward, with these continuous agents, the quantization is done with gradient-based.
Different from Ref.~\refcite{rq} but similar to PACT, LSQ~\cite{lsq} introduces learnable parameters in the quantizer. But the new parameter is for learning step size instead of clipping value. Eventually, they made the 3-bit models reach the full precision baseline accuracy. 
Like LSQ, Ref.~\refcite{stquantization} is also for learning quantization interval, but it utilized the power of the attention mechanism and it is realized with the squeeze and threshold (ST) block. Recently, Ref.~\refcite{ddq} presented differentiable dynamic quantization (DDQ), a fully differentiable technique to learning bitwidth, clipping value, and quantization interval.

\section{Our Approach}

Squeeze-and-threshold (ST) block was proposed in {Ref.~\refcite{stquantization}} to learn the activation quantization threshold. It utilizes the information of activations itself to quantize the activations. Moreover, the real-value ST branch bypasses the quantization step and can fully back-propagate the gradient to the previous layers. However, the activation quantization in ST is input channel-wise and output channel-wise. It is not consistent with the TensorFlow-Lite~\cite{tensorflow} or PyTorch~\cite{pytorch} quantization, in which the activation quantization is only output channel-wise. The difference makes the network inference not so efficient on currently available hardware.
Therefore, in this work, we propose an improved version. Its final activation quantization is only output channel-wise. Furthermore, we extend the ST block to be double stages and further improve the performance of the quantized networks. In the following content of this section, we will detail the improved method, the double-stage ST. For ease of illustration, we adapt the following notations: The tensor in the expression is in bold. Notation $H$, $W$ and $C$ represent the height, width, and channel, respectively, while $N$ could be the batch of activation or number of filters. Lastly, notation $\mathbb{B}$ is the bit width. 

\subsection{Activation Quantization}
The integration of double-stage-ST-based activation quantization into a network is shown in Fig.~\ref{factiquan}. The activation quantization corresponds to the part marked with green and blue color, named $\mathbf{ActiQuan}$. With our quantization solution, the network architecture stays the same as before quantization. 

\begin{figurehere}
\begin{center}
\includegraphics[width=3in]{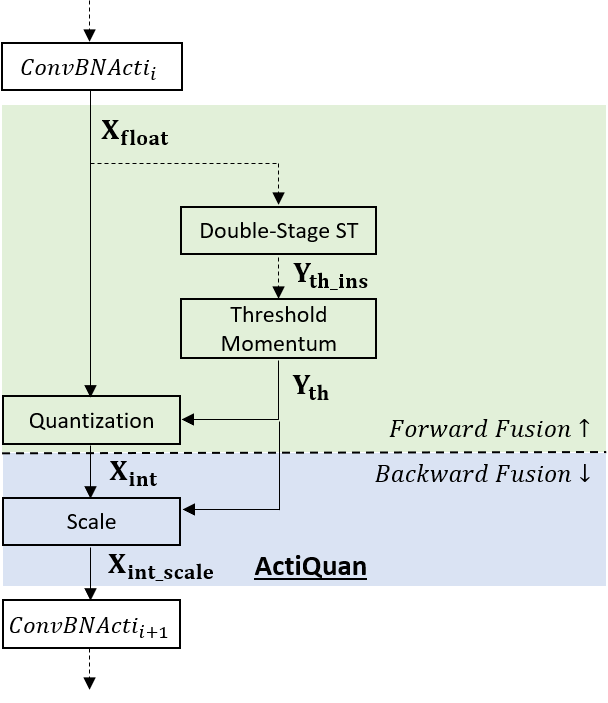}
\caption{Integration of double-stage-ST-based activation quantization, $\mathbf{ActiQuan}$, into networks. $\mathbf{X_{float}}$ is the full-precision activation from previous layer with a shape of $[N, H, W, C]$. $\mathbf{Y_{th\_ins}}$ is the instant activation quantization threshold with a shape of $[1, 1, 1, C]$. $\mathbf{Y_{th}}$ has the same shape as $\mathbf{Y_{th\_ins}}$ and is iteratively updated by $\mathbf{Y_{th\_ins}}$. $\mathbf{X_{int}}$ is the integer activation having $2^{\mathbb{B}}$ kinds of values. $\mathbf{X_{int_scale}}$ is the full-precision activation but with only $2^{\mathbb{B}}$ options. }
\label{factiquan}
\end{center}
\end{figurehere}

The double-stage-ST block extracts the instant threshold $\mathbf{Y_{th\_ins}}$ from each batch input activations. Then $\mathbf{Y_{th\_ins}}$ iteratively updates the final threshold $\mathbf{Y_{th}}$ with 
\begin{equation*}
\mathbf{Y_{th}^{+}} = (1 - \mathbb{M})\times \mathbf{Y_{th}^{-}} + \mathbb{M}\times \mathbf{Y_{th\_ins}},
\label{ethupdate}
\end{equation*}
where $\mathbf{Y_{th}^{+}}$ and $\mathbf{Y_{th}^{-}}$ are the newly updated and previous threshold, respectively. $\mathbb{M}$ is the threshold update coefficient and changes along with training,
\begin{equation*}
\mathbb{M} = \mathbb{M}_{min}\times(1 - \cos{(\frac{\mathbb{E}_{cur}}{\mathbb{E}_{tol}})}) + \cos{(\frac{\mathbb{E}_{cur}}{\mathbb{E}_{tol}})},
\label{emomentum}
\end{equation*}
where $\mathbb{M}_{min}$ is a hyperparameter, the minimum momentum factor. $\mathbb{E}_{cur}$ and $\mathbb{E}_{tol}$ are the current training epoch and total training epochs, respectively.

In quantization, each channel of activation is quantized with the corresponding threshold in $\mathbf{Y_{th}}$,
\begin{equation*}
\mathbf{X_{int}} = 
\begin{cases} 
    \mathrm{sign}(\mathbf{X_{float}}) * 0.5 + 0.5, & \mbox{if } \mathbb{B} = 1 \\ 
    \mathrm{clip}(\mathrm{round}(\frac{\mathbf{X_{float}}}{\mathbf{Y_{th}}}), \mathbb{V^-}, \mathbb{V^+}), & \mbox{if } \mathbb{B} > 1
\end{cases}
\end{equation*}
where $\mathrm{sign(\cdot)}$ and $\mathrm{round(\cdot)}$ are the signum and rounding functions, respectively. For multi-bit quantization, $\mathbf{X_{int}}$ is clipped in the range of $[\mathbb{V^-}, \mathbb{V^+}]$ by the clipping function $\mathrm{clip(\cdot)}$. For pure positive activation, $\mathbb{V^-}=0$ and $\mathbb{V^+} = 2^{\mathbb{B}} - 1$, otherwise, $\mathbb{V^-} = -2^{\mathbb{B}-1}$ and $\mathbb{V^+} = 2^{\mathbb{B}-1} - 1$.  In the backward path, the non-differentiable functions $y=\mathrm{sign}(x)$ and $y=\mathrm{round}(x)$ are approximated by $y=\mathrm{sigmoid}(4x)$ and $y=x$, respectively.

To maintain the activation statistics and still limit its value with only $2^{\mathbb{B}}$ possibilities.
\begin{equation*}
\mathbf{X_{int\_scale}} = \mathrm{mean}(\mathbf{Y_{th}})\times \mathbf{X_{int}},
\label{eactiintscale}
\end{equation*} 
where $\mathrm{mean}(\cdot)$ obtains the mean value of the input tensor.
With this setting, the final quantization is compatible with the quantization flow in TensorFlow and PyTorch and is also friendly for currently available hardware.

\subsection{Double-Stage Squeeze-and-Threshold}
Fig.~\ref{fdoublest} shows the construction of the double-stage squeeze-and-threshold (ST) block, which is for learning instant activation quantization threshold. 
\begin{figurehere}
\begin{center}
\includegraphics[width=1.25in]{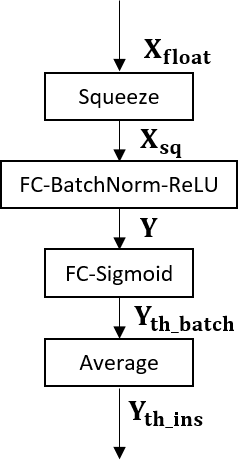}
\caption{Double-Stage squeeze-and-Threshold block.}
\label{fdoublest}
\end{center}
\end{figurehere}

The squeeze operation is done with a global average operation, generating $\mathbf{X}_{sq}\in\mathbb{R}^{N\times 1 \times 1\times C}$. Compared with {ST\cite{stquantization}}, an additional fully-connected-batch-normalization-ReLU layer (denoted as FC-BatchNorm-ReLU in Fig.~\ref{fdoublest}) is added to increase the non-linearity, producing $\mathbf{Y}\in\mathbb{R}^{N\times 1 \times 1\times C}$. Subsequently, $\mathbf{Y}$ is transferred to activation quantization threshold $\mathbf{X}_{th\_batch}\in\mathbb{R}^{N\times 1 \times 1\times C}$ by a fully-connected-sigmoid layer. The final average operation computes the arithmetic mean over $\mathbf{X}_{th\_batch}$ along the batch axis, generating the channel-wise activation threshold, $\mathbf{Y}_{th\_ins}\in\mathbb{R}^{1\times 1 \times 1\times C}$.

\subsection{Inference Optimization}
The data only flows through the double-stage ST branch in the training phase. In the inference phase, the iteratively updated $\mathbf{Y_{th}}$ will be used instead. Therefore, double-stage ST block will not introduce computation overhead during inference. In this section, we illustrate how to reconstruct the parameters and operations of $\mathbf{ActiQuan}$.

In {TensorFlow-Lite\cite{tensorflow}}, the quantized convolution-batch-normalization-relu layer will be optimized into
\begin{equation}
\begin{split}
\mathbf{Y_{bn}}&=\mathbf{S}(\mathbf{Y_{conv\_int}} + \mathbf{B_{int}})\\
\mathbf{Y_{relu\_int}}&=\mathrm{clip}(\mathrm{round}(\mathbf{Y_{bn}}), \mathrm{MIN}, \mathrm{MAX}),
\label{eybn}
\end{split}
\end{equation}
where $\mathbf{Y_{conv\_int}}\in\mathbb{Z}^{N\times H\times W\times C}$,  $\mathbf{Y_{bn}}\in\mathbb{R}^{1\times 1\times 1\times C}$ and $\mathbf{Y_{relu\_int}}\in\mathbb{Z}^{N\times H\times W\times C}$ are the output of convolution, batch normalization, and activation function, respectively. $\mathbf{B_{int}}\in\mathbb{Z}^{1\times 1\times 1\times C}$ is the bias. $\mathbf{S}=\frac{\mathbf{S_w}S_x^i}{S_x^o}\in\mathbb{R}^{1\times 1\times 1\times C}$, in which $\mathbf{S_w}\in\mathbb{R}^{1\times 1\times 1\times C}$ is the weight scaling factor, $S_x^i$ is the scalar scaling factor for the input activation and $S_x^o$ is the scalar re-quantization scaling factor for the output activation.

Looking back at our quantization, since ReLU executes maximum operation $\mathrm{max}(0, x)$ and the threshold $\mathbf{Y_{th}}$ used in the quantization step is non-negative, the green part marked as \emph{Forward Fusion} in Fig.~\ref{factiquan} can be fused into the previous batch normalization layer. In addition, the nonlinearity function is no longer required. For the blue part marked as \emph{Backward Fusion}, because of the linearity of convolution operation, its scalar scaling parameter, $\mathrm{mean}(\mathbf{Y_{th}^i})$, can be fused into the following batch-normalization layer. Eventually, for our quantization, Eq.~(\ref{eybn}) becomes
\begin{equation*}
\begin{split}
\mathbf{Y_{bn}}&=\mathbf{S_{th}}(\mathbf{Y_{conv\_int}} + \mathbf{B_{int}})\\
\mathbf{Y_{relu\_int}}&=\mathrm{clip}(\mathrm{round}(\mathbf{Y_{bn}}), \mathbb{V^-}, \mathbb{V^+}),
\label{eybnfuse}
\end{split}
\end{equation*}
where $\mathbf{S_{th}}=\alpha\frac{\mathbf{S_w}}{\mathbf{Y_{th}^o}}\in\mathbb{R}^{1\times 1\times 1\times C}$, $\mathbf{Y_{th}^o}\in\mathbb{R}^{1\times 1\times 1\times C}$ is the activation threshold of the output activation and $\alpha=\mathrm{mean}(\mathbf{Y_{th}^i})$ is a scalar value and equals to mean of the input activation threshold.

\section{Experiments}
In this section, we prove the feasibility of the double-stage-ST-based activation quantization and perform comparisons with previous works. We quantize both activation and weight with the same bit width. The weight quantization uses the method proposed by {LSQ~\cite{lsq}}. Same as previous {work~\cite{binarydenseNet,reactnet,lsq,pact}}, the first and last layers keep full precision. All the experiments are done on dataset {ImageNet\cite{imagenet}}.

\subsection{Improvement of Double-Stage Squeeze-and-Threshold}
In this subsection, we execute a comparison with previous work ST\cite{stquantization} and show the improvement. To have a fair comparison, we use the same full precision model of ResNet18 as in ST, which is available on the PyTorch official website and can achieve a TOP1/TOP5 accuracy of $69.7\%$/$89.1\%$. All of the training hyper-parameters are the same as ST, except the optimizer learning rate and weight decay. The comparison result is demonstrated in Table~\ref{tab1}. As the result shows, compared to ST, our proposed double-stage ST improves the TOP1 accuracy of 1-bit, 2-bit, 3-bit, and 4-bit models by $0.2\%$, $0.4\%$, $1.0\%$, and $1.2\%$, respectively. Moreover, the accuracy of the 3-bit and 4-bit models of double-stage ST are beyond the full-precision accuracy. The detailed training settings are available in Appendix~\ref{appresnet1}.
\begin{table*}
\tbl{Performance comparison between ST and double-stage ST. Comparison is evaluated by TOP1/TOP5 accuracy in percentage on ImageNet. The full-precision model ResNet-18 TOP1/TOP5 accuracy is $69.7\%$/$89.1\%$.  \label{tab1}}
{\begin{tabular}{@{}lcccc@{}}
\toprule
Method & 1-bit & 2-bit & 3-bit & 4-bit\\ \colrule
ST~\cite{stquantization}        & 57.5/80.4 & 66.7/86.9 & 68.8/88.5 & 69.5/88.9 \\
\textbf{Double-Stage ST (Ours)} & \bfseries57.9/80.6 & \bfseries67.1/87.3 & \bfseries69.8/89.1 & \bfseries70.7/89.6\\
\botrule
\end{tabular}}
\end{table*}

\subsection{Comparison with state-of-the-art methods}
In this subsection, we compare our double-stage ST method with state-of-the-art methods. The comparison result is available in Table~\ref{tab2}. All the TOP1/TOP5 accuracy of other methods are from the original paper.

For the binary model, the comparison is executed on the ImageNet\cite{imagenet} dataset and ReActNet\cite{reactnet} model. We use our double-stage-ST-based activation binarization to replace ReActNet's original activation binarization. Same as the original work of {ReActNet\cite{reactnet}}, we train the network with two steps. Only activation is binarized in the first step, and weight is precisely trained. In the second step, the model is initialized with step one model. Then both activation and weight are binarized. The detailed training settings are availalbe in Appendix~\ref{appreactnet}.

For the multi-bit models, the comparison is performed on the ImageNet\cite{imagenet} dataset and the ResNet-18\cite{resnet} model. Since most state-of-the-art multi-bit quantization methods use a full-precision ResNet-18 with a TOP1/TOP5 accuracy of $70.5\%$/$89.8\%$, for a fair comparison, we re-train ResNet-18 to achieve such accuracy and use it as the initial model for quantization. To achieve the best accuracy, we train the network with two steps as training ReActNet. The detailed training settings are in Appendix~\ref{appresnet2}. 

\begin{table*}
\tbl{Performance comparison between double-stage ST and other state-of-art methods on ReActNet and ResNet-18. Comparison is evaluated by TOP1/TOP5 accuracy in percentage on ImageNet. ReActNet is trained from scratch and ResNet-18 quantization starts with a full-precision model with TOP1/TOP5 accuracy of $70.5\%$/$89.8\%$. \label{tab2}}
{\begin{tabular}{@{}lcccc@{}}
\toprule
                        &\multicolumn{3}{c}{ResNet-18}                              &\multicolumn{1}{c}{ReActNet}   \\ \colrule
Method                  &2-bit              &3-bit              &4-bit              &1-bit                          \\ \colrule
\textbf{Double-Stage ST (Ours)} &\bfseries68.1/87.9 &\bfseries70.9/89.6 &\bfseries71.8/90.0 &\bfseries70.7/89.3     \\
DDQ(2021)\cite{ddq}     &---/---            &---/---            &71.1/---           &---/---                        \\
HAWQV3(2021)\cite{hawqv3}&---/---           &68.5/---           &---/---            &---/---                        \\
LSQ(2020)\cite{lsq}     &67.9/88.1          &70.6/89.7          &71.2/90.1          &---/---                        \\
DSQ(2019)\cite{dsq}     &65.2/---           &68.7/---           &69.6/---           &---/---                        \\
LQ-Nets(2018)\cite{lqnet}&64.9/85.9         &68.2/87.9          &69.3/88.8          &---/---                        \\
NICE(2018)\cite{nice}   &---/---            &67.7/88.2          &69.8/89.2          &---/---                        \\
PACT(2018)\cite{pact}   &64.4/---           &68.1/---           &69.2/---           &---/---                        \\
ReActNetAdam(2021)\cite{reactnetadam}&---/--- &---/---          &---/---            &70.5/89.1                      \\
ReActNet(2020)\cite{reactnet}&---/---       &---/---            &---/---            &69.4/88.6                      \\
\botrule
\end{tabular}}
\end{table*}

\section{Conclusion}
In the actual application of ST quantification, we found that its input-channel-wise and output-channel-wise quantization weakened the efficiency of the quantization network and brought difficulties to deployment. As a result, we further investigate the attention mechanism and upgraded it to a two-stage ST. In the end, the two-stage ST quantization performs better than other competitors and demonstrates state-of-the-art results.

\nonumsection{Acknowledgments} 
This work is funded by the German Research Foundation (DFG, Deutsche Forschungsgemeinschaft) as part of Germany’s Excellence Strategy – EXC 2050/1 – Project ID 390696704 – Cluster of Excellence “Centre for Tactile Internet with Human-in-the-Loop” (CeTI) of Technische Universität Dresden. 

We are grateful to the Centre for Information Services and High Performance Computing [Zentrum für Informationsdienste und Hochleistungsrechnen (ZIH)] TU Dresden for providing its facilities for high throughput calculations. 

\appendix{Training setting for ResNet-18 binarization and quantization in Table~\ref{tab1}}
\label{appresnet1}
In this appendix, the training settings of models in Table~\ref{tab1} are listed in detail. The models were trained for 90 epochs with a batch size of 256 on two NVIDIA A100 GPUs. $\mathbb{M}_{min}$ is set to 0.1. The learning rate is set using a cosine annealing schedule with a minimum learning rate of $10^{-4}$. Stochastic gradient descent (SGD) with momentum 0.9 and Nesterov momentum is used for optimization. 
For the newly introduced operation, their trainable parameters are initialized with the default one in PyTorch. For the non-trainable parameter, the activation threshold is set to zero at the beginning. The weight quantization solution stays the same as ST\cite{stquantization}.
The optimizer initial learning rate $\gamma$ for 1-bit, 2-bit, 3-bit, and 4-bit models are $0.125$, $0.05$, $0.035$, and $0.025$, respectively. Whereas the optimizer weight decay $\lambda$ for 1-bit, 2-bit, 3-bit, and 4-bit models are $1\times 10^{-6}$, $3\times 10^{-5}$, $3\times 10^{-5}$, and $3\times 10^{-5}$, respectively. Whereas the optimizer weight decay $\lambda$ is $1\times 10^{-6}$ for binarization and $3\times 10^{-5}$ for multi-bit quantization.

\appendix{Training setting for ReActNet in Table~\ref{tab2}}
\label{appreactnet}
For the newly introduced hyperparameter $\mathbb{M}_{min}$, we set $\mathbb{M}_{min}=0.1$. ResNet-101 is used as the teacher model. All of the training hyper-parameters are the same as original {ReActNet\cite{reactnetadam}}, except the optimizer learning rate and weight decay shown in Table~\ref{tabsettingc}. The initial learning rate $\gamma$ and weight decay $\lambda$ are $0.75\times 10^{-3}$ and $1\times 10^{-6}$ in the first step, and $0.5\times 10^{-3}$ and $0.0$ in the next step.

\appendix{Training setting for ResNet-18 binarization and quantization in Table~\ref{tab2}}
\label{appresnet2}
In this appendix, the training settings of models in Table~\ref{tab2} are listed in detail.
For the ResNet-18 quantized model, we employ two-step training. In the first step, the model is initialized with the full-precision model with TOP1/TOP5 accuracy of $70.5\%$/$89.8\%$ and only activation is quantized. In the second step, the model is initialized with the model from step one. Both activation and weight are quantized. Furthermore, we apply labeling smoothing\cite{labelsmoothing} with a factor of $0.1$. 
Table~\ref{tabsettingb} shows the SGD optimizer's initial learning rate $\gamma$ and weight decay $\lambda$.
The other training settings are the same as in Appendix~\ref{appresnet1}.
\begin{tablehere}
\tbl{Initial learning rate $\gamma$ and weight decay $\lambda$ settings for ResNet-18 quantization. The precision "A\emph{x}W\emph{y}" represents \emph{x}-bit activation and \emph{y}-bit weight.\label{tabsettingb}}
{\begin{tabular}{@{}lcccc@{}}
\toprule
Model                       &step   &precision  &$\gamma$   &$\lambda$ \\ 
\colrule
\multirow{2}{3em}{2-bit}    &step1  &A2W32      &0.1        &$4\times 10^{-6}$\\ 
                            &step2  &A2W2       &0.05       &$4\times 10^{-6}$\\
\colrule
\multirow{2}{3em}{3-bit}    &step1  &A3W32      &0.035      &$7\times 10^{-6}$\\ 
                            &step2  &A3W3       &0.035      &$7\times 10^{-6}$\\
\colrule
\multirow{2}{3em}{4-bit}    &step1  &A4W32      &0.025      &$1\times 10^{-5}$\\ 
                            &step2  &A4W4       &0.025      &$1\times 10^{-5}$\\
\botrule
\end{tabular}}
\end{tablehere}

The achieved TOP1/TOP5 accuracy of each step of the ResNet-18 quantized models are shown in Table~\ref{tabsettingc}
\begin{tablehere}
\tbl{TOP1/TOP5 accuracy of each step of the ResNet-18 quantized models. The corresponding full-precision ResNet-18 has a TOP1/TOP5 accuracy of $70.5/89.8$. \label{tabsettingc}}
{\begin{tabular}{@{}lccc@{}}
\toprule
model                       &step   &precision       &TOP1/TOP5  \\ 
\colrule
\multirow{2}{3em}{2-bit}    &step1  &A2W32           &69.162/88.386          \\ 
                            &step2  &A2W2            &68.114/87.884          \\
\colrule
\multirow{2}{3em}{3-bit}    &step1  &A3W32           &71.342/89.848          \\ 
                            &step2  &A3W3            &70.906/89.560          \\
\colrule
\multirow{2}{3em}{4-bit}    &step1  &A4W32           &71.804/90.192          \\ 
                            &step2  &A4W4            &71.822/89.988         \\
\botrule
\end{tabular}}
\end{tablehere}

\bibliographystyle{ws-ijns}
\bibliography{ref}

\end{multicols}
\end{document}